\pdfoutput=1
\documentclass{article}

\usepackage[final]{corl_2020} 

\usepackage{amsmath}
\usepackage{multirow}
\usepackage{graphicx}
\usepackage{subcaption}
\usepackage{amsfonts}
\newcommand{\etal}{\textit{et al}.~}
\newcommand{\ie}{\textit{i}.\textit{e}.}
\newcommand{\eg}{\textit{e}.\textit{g}.}
\newcommand{\etc}{\textit{etc}.~}
\usepackage{array}
\usepackage{booktabs,caption}
\newcolumntype{P}[1]{>{\centering\arraybackslash}p{#1}}
\newcolumntype{M}[1]{>{\centering\arraybackslash}m{#1}}
\usepackage{bm}
\usepackage{makecell}

\title{SelfVoxeLO: Self-supervised LiDAR Odometry with Voxel-based Deep Neural Networks }

\author{
\makecell[c]{Yan Xu$^1$~~~Zhaoyang Huang$^{1,2}$~~~Kwan-Yee Lin$^{1,3}$~~~Xinge Zhu$^1$~~~Jianping Shi$^3$\\ Hujun Bao$^2$~~~Guofeng Zhang$^2$~~~Hongsheng Li$^1$}
	\\
\makecell[c]{$^{1}$Multimedia Laboratory, The Chinese University of Hong Kong\\$^{2}$State Key Lab of CAD\&CG, Zhejiang University~~~ $^{3}$SenseTime Research}\\
}

\begin{document}
\maketitle

\begin{abstract}
  Recent learning-based LiDAR odometry methods have demonstrated their competitiveness. 
  However, most methods still face two substantial challenges: 1) the 2D projection representation of LiDAR data cannot effectively encode 3D structures from the point clouds; 
  2) the needs for a large amount of labeled data for training limit the application scope of these methods.
    In this paper, we propose a self-supervised LiDAR odometry method, dubbed SelfVoxeLO, to tackle these two difficulties.
    Specifically, we propose a 3D convolution network to process the raw LiDAR data directly, which extracts features that better encode the 3D geometric patterns. 
    To suit our network to self-supervised learning, 
    we design several novel loss functions that 
    utilize the inherent properties of LiDAR point clouds. 
    Moreover, an uncertainty-aware mechanism is incorporated in the loss functions to alleviate the interference of moving objects/noises. 
   We evaluate our method's performances on two large-scale datasets, \ie, KITTI and Apollo-SouthBay.
   Our method outperforms 
   state-of-the-art unsupervised methods by 27\%/32\% in terms of translational/rotational errors on the KITTI dataset 
   and also performs well on the Apollo-SouthBay dataset. By including more unlabelled training data, our method can further improve performance comparable to the supervised methods.

  \end{abstract}
  


\section{Introduction}

Ego-motion estimation from temporal sequences of sensor data, also known as odometry, is of fundamental importance for many robotic vision tasks, including navigation, mapping, virtual/augmented reality, \etc 
Compared with visual sensors, the LiDAR can capture richer 3D geometric information of the environments and is robust against varying lighting conditions. Hence, a reliable LiDAR odometry system is desirable for localization systems.

\begin{figure*}
    \centering
    \begin{subfigure}[t]{0.6\textwidth}
    \centering
    \includegraphics[width=1\linewidth, trim=1cm 10cm 7cm 1cm, clip]{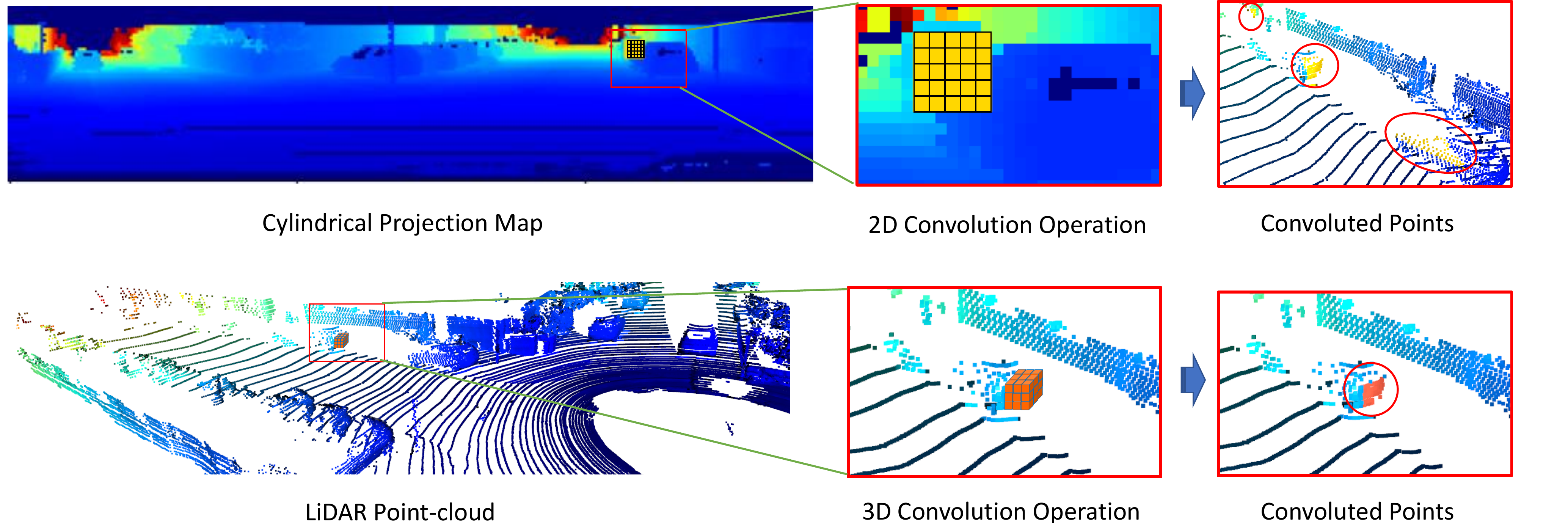}
    \caption{}
    \label{fig:3d_conv}
    \end{subfigure}
    \hfil
    \begin{subfigure}[t]{0.32\textwidth}
        \includegraphics[width=\linewidth, trim=0.9cm 12.5cm 31cm 1cm, clip]{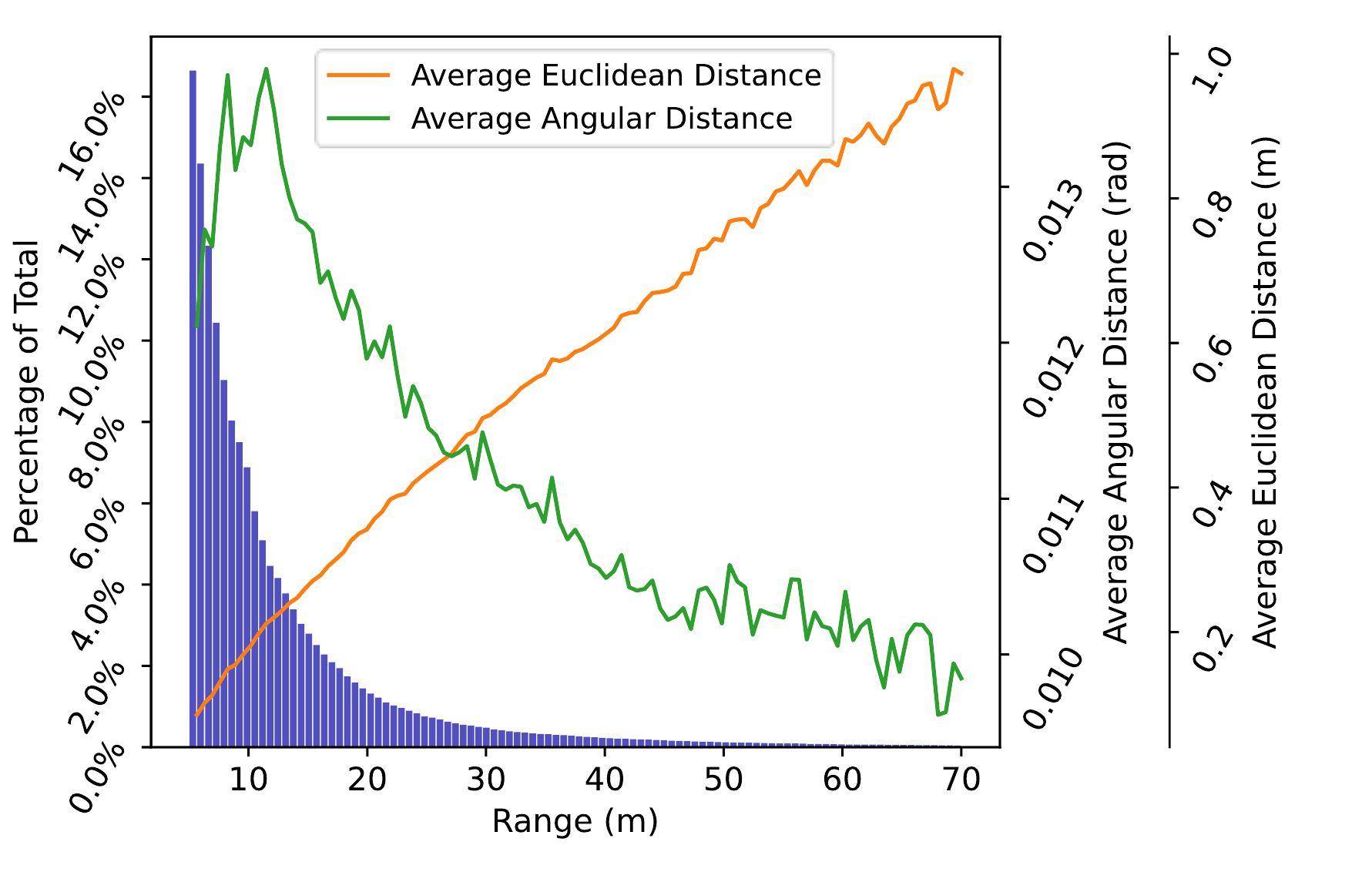}
    \caption{}
    \label{fig:distribution}
    \end{subfigure}
    \caption{(a) The difference between the 2D convolution and the 3D convolution. (b) The distribution of point clouds with respect to LiDAR range value, 
    and the measured average distances with different metrics (\ie, Euclidean distance and angular distance~\cite{angerr}) in each bin.
    Large Euclidean distances in distant regions make the unreliable remote points count more, which is not desirable. 
    }
    \vspace{-4ex}

\end{figure*}

The classic methods~\cite{arun1987least,segal2009generalized,serafin2015nicp,zhang2014loam,shan2018lego} are mainly based on the point registration and work well in ideal scenarios, but they might fail in practice due to the sparse nature of point clouds and environmental noises.
Typically, ICP~\cite{arun1987least} and its variants~\cite{arun1987least,segal2009generalized,serafin2015nicp} iteratively find the point correspondences with nearest-neighbor searching and optimize for the pose transformations.
This optimization procedure ignoring the correspondence reliability can easily run into local optimum especially when noise and dynamic objects exist. 
In the past a few years, the advances in deep learning have significantly advanced state of the arts in odometry estimation. 
Seminal works~\cite{konda2015learning,wang2017deepvo,zhou2017unsupervised}, demonstrate the feasibility of 6-DOF pose regression via convolutional neural network for visual odometry estimation. Following the visual pipelines, several LiDAR odometry estimation approaches have been proposed~\cite{velas2018cnn,li2019net,chounsupervised}, where they often project the 3D LiDAR points onto a cylindrical surface and then adopt the similar framework from CNN-based visual odometry methods. 
However, projecting the point clouds to the cylindrical surface inevitably alters the 3D topology and cannot effectively capture the geometric information. As illustrated in Fig.~\ref{fig:3d_conv}, the 2D convolution  on the cylindrical projection map might process the points on objects far away from each other, which 
ignore the 3D topology relations and contaminate the encoded features.
A straightforward idea to tackle the issue of cylindrical projection is to adopt 3D representations and process the points in the 3D space directly.
Therefore, the 3D convolution network is an appealing alternative. In contrast to the 2D convolution, the 3D convolution can better retain the 3D topology relations and structures during the hierarchical feature extraction. 
On the other hand, it should be noted that the significant progress of CNN-based methods relies, to a large extent, on large-scale annotated training data, which is not always feasible in practice, due to the huge cost needed for large-scale annotations.
How to train a LiDAR odometry network in an unsupervised manner is still an imperative problem.

In this work, we develop an self-supervised learning-based LiDAR odometry method with 3D convolution networks.
Specifically, our network first voxelizes the point clouds into fine-grid voxel cells, and extracts 3D features via 3D convolutional neural networks.
Then, the extracted features are fed into our odometry regression network 
to predict the final 6-DOF ego-motions. 
To suit our network to self-supervised training, we analyse the inherent properties of the LiDAR point clouds and propose several losses: the spherical reprojection loss that essentially pushes the network to focus on the nearby stable points, the transformation residual loss to stabilize the training, and the deep flow supervision loss to facilitate the point-wise feature learning.  
Furthermore, to mitigate the interference from noise and dynamic objects, we also propose to estimate the correspondence-pair confidences, which are incorporated into our self-supervised losses to set lower weights for the unreliable point correspondences. 

Our contributions can be summarized as follows:
~(1)~
We abandon the common 2D projection-based LiDAR odometry framework 
and investigate the effectiveness of voxel-based 3D geometric representation.
We propose a framework that predicts ego-motions from raw LiDAR points based on 3D convolutional neural networks, which are trained in an unsupervised manner. 
~(2)~Several self-supervised loss functions are introduced, \ie, the spherical reprojection loss, the range alignment loss, the transformation residual  loss and the deep flow supervision loss
   to train ego-motion prediciton without any annotations.
  Furthermore, we incorporate an uncertainty-aware mechanism into the loss functions to mitigate the interference of moving objects and noise.
~(3)~The proposed method achieves state-of-the-art performances on two public odometry datasets, \ie, the KITTI dataset~\cite{geiger2012we} and the Apollo-SouthBay Dataset~\cite{lu2019l3}.

\vspace{-1ex}
\section{Related Work}
\vspace{-2ex}
\textbf{Odometry.} 
LiDAR odometry algorithms can be categorized as two-frame methods~\cite{arun1987least,segal2009generalized,serafin2015nicp} and multi-frame methods~\cite{zhang2014loam,shan2018lego}.
The classic two-frame methods are mostly based on the point registration, where ICP~\cite{arun1987least} and its variants~\cite{rusinkiewicz2001efficient,segal2009generalized,serafin2015nicp,pathak2010fast} are typical exemplars. 
The ICP iteratively finds the point correspondences and optimizes for the pose transformation between two LiDAR pointclouds util convergence. 
Different ICP methods may apply different weights on the correspondence pairs during optimization 
according to the geometric characteristics. 
However they often fail to model the moving objects that 
violate the algorithm assumption.
Moreover, most of these methods are too computationally expensive to be applied in real-time systems. 
The multi-frame algorithms (often referred as mapping~\cite{zhang2014loam,shan2018lego,behley2018efficient}) are often used to refine the two-frame based estimation by incorporating more frames into optimization. 
They are computationally heavier and usually runs in the backend at a lower frequency.   
In our work, we mainly focus on the two-frame method which is more fundamental and can be combined with multi-view methods. 
Recently, many CNN-based odometry methods were proposed. 
Initially, several seminal methods were proposed for visual odometry~\cite{konda2015learning,wang2017deepvo,zhou2017unsupervised}. 
More recently, researchers used CNN in LiDAR odometry~\cite{li2019net,velas2018cnn}. They represented the LiDAR pointclouds by cylindrical projection 
and then borrowed the network architectures from the visual odometry methods. 
Cho \etal~\cite{chounsupervised} extended this pipeline to unsupervised learning inspired by the unsupervised visual odometry~\cite{zhou2017unsupervised,li2018undeepvo}. 
{However, the cylindrical projection may lose the spatially geometric structures of the input pointclouds, which leads to unreliable feature extraction. 
} 
\\
\textbf{3D network.} In the past few years, various 3D DNNs have been proposed to better handle the 3D data. 
The 3D convolution operation is also applied on local areas centered at fixed grids, but the operation are conducted in the 3D space rather than the 2D space. 
{PointNet~\cite{qi2017pointnet,qi2017pointnet++} determines the grid locations with furthest sampling and find the related feature points in respective local areas with nearest-neighbor searching.} 
It achieves high flexibility but sacrifices speed.    
To achieve high efficiency, Zhou \etal~\cite{zhou2018voxelnet} proposed to split the space into voxels uniformly and apply the 3D convolution on these uniform 3D grids. 
Gu \etal~\cite{gu2019hplflownet} proposed the 3D convolution on permutohedral lattices. 
Hanocka \etal~\cite{hanocka2019meshcnn} proposed a new set of convolution operations to handle 3D mesh data. 
Compared with 2D convolution, 3D convolution is more suitable to process the 3D data and has made great success in 3D object detection~\cite{zhou2018voxelnet,yan2018second,lang2019pointpillars,qi2018frustum,you2019pseudo}, 3D scene flow predicition\cite{gu2019hplflownet,liu2019flownet3d} and 3D semantic segmentaion \cite{qi2017pointnet,qi2017pointnet++,hanocka2019meshcnn,prokudin2019efficient}.
However, few investigation has been made using 3D convolutional networks on LiDAR odometry.	

\vspace{-1ex}
\section{Method}
\vspace{-2ex}
\begin{figure*}
    \centering
    \includegraphics[width=0.87\linewidth, trim=0cm 4cm 5cm 0.5cm, clip]{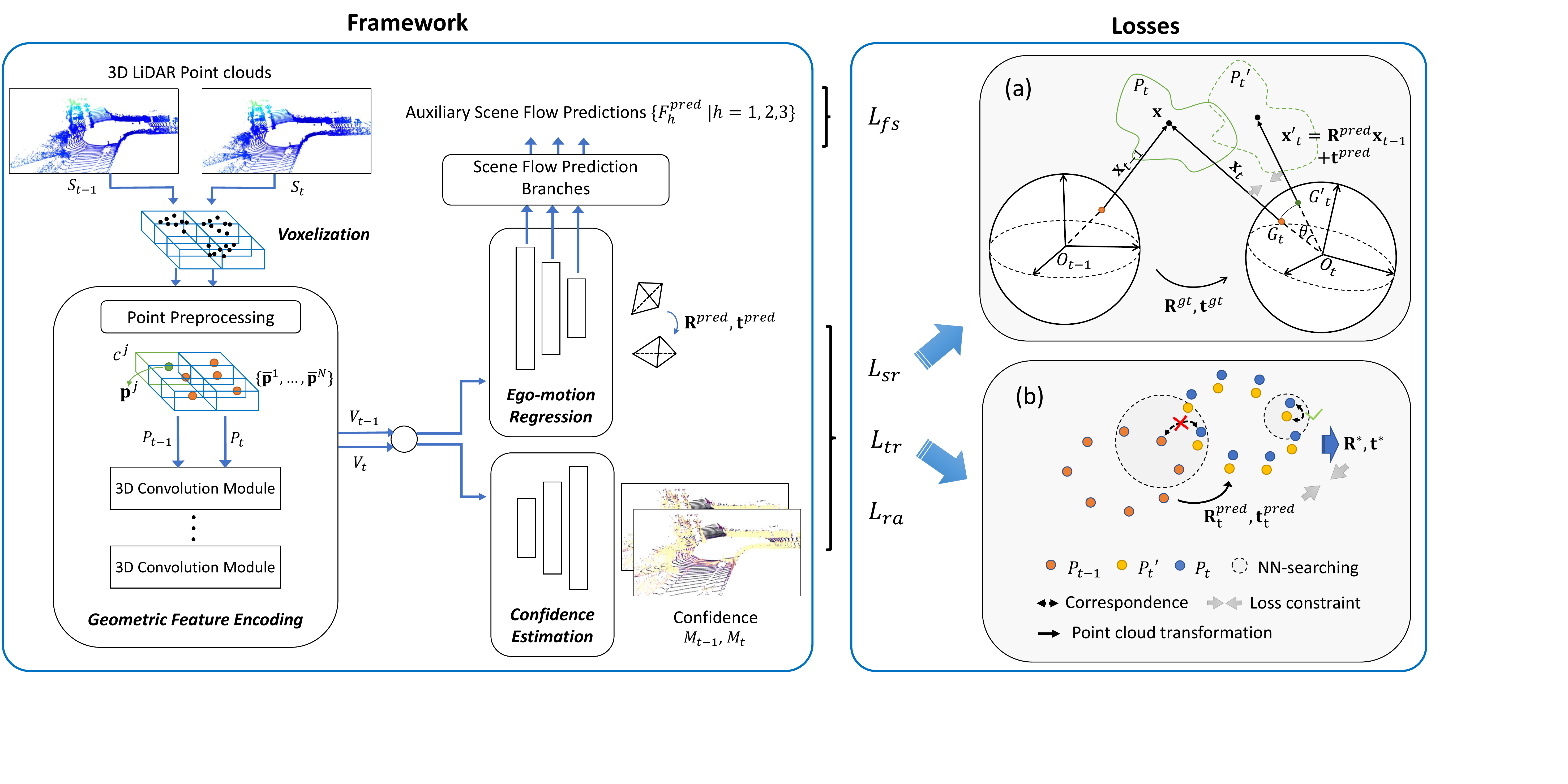}
    \caption{
      Framework overview and loss functions. 
    The input LiDAR point clouds 
     are voxelized and then fed to the 3D \textit{\textbf{geometric feature encoding}} module, where the points are pre-downsampled 
    (obtaining point sets $P_t$ and $P_{t-1}$)
     and encoded by the 3D convolution modules to obtain the 3D geometric feature sets $V_{t-1}$ and $V_{t}$.
     Based on these 3D features, the \textit{\textbf{ego-motion regression}} module and the \textit{\textbf{confidence estimation}} module predict the ego-motion $(\mathbf{R}^{pred}, \mathbf{t}^{pred})$ and the voxel-wise confidence estimations ($M_{t-1}$ and $M_{t}$)  
     for each sweep.
      The sub-figure (a) and (b) demonstrate our spherical reprojection loss and transformation residual loss, which are elaborated in Sec.~\ref{sec:odom_regress}.
    }
    \label{fig:framework}
    \vspace{-3ex}
\end{figure*}

To tackle the mentioned challenges, 
\ie, the limitation of 2D convolution and the need of large-scale labeled data,
our framework aims to estimate the 6-DOF agent's pose transformation 
$(\mathbf{R}^{pred},\mathbf{t}^{pred})$ 
from two successively scanned point clouds $S_{t-1}$ and  $S_{t}$, \emph{without ground-truth transformations}.
Different from the previous works~\cite{li2019net,velas2018cnn,chounsupervised} adopting 2D convolutions on cylindrical projection maps of the point clouds, we use fine-grid voxels to represent the LiDAR point cloud and propose to directly process the point clouds with 3D convolutions,   
which can better maintain the 3D geometric information than the 2D convolutions in existing projection-based methods. 
To enable stable training of LiDAR-based odometry without ground-truth,
we introduce three self-supervised losses and uncertainty-aware mechanism.
As shown in Fig.~\ref{fig:framework},
our framework mainly consists of three modules: (1) 3D geometric feature encoding, (2) ego-motion regression and 
(3) confidence estimation. 
We will detail each part and the proposed unsupervised loss functions in the ensuing sections. 

\subsection{3D Geometric Feature Encoding}
\vspace{-1ex}
The odometry task requires real-time response as well as high precision. 
As mentioned above, the 3D convolution on 3D feature spaces can efficiently preserve the geometric information from the scene, which is an appealing candidate. 
However, the 3D space is unbounded with low data occupancy. Directly applying 3D convolutions to 3D point clouds is resource intensive.
Previous works make a compromise and project the LiDAR point clouds to cylindrical maps to persist with the legacy of 2D CNNs, which obviously is not an optimal solution.    
Recently, with the development of 3D CNNs, many efficient 3D operators spring up making efficient 3D geometric feature encoding possible. 
Hence, to achieve precise feature encoding while maintaining real-time speed, we conduct 3D feature encoding in our framework by adopting the voxel representation and the recent proposed sparse submanifold convolutions~\cite{graham2017submanifold}.

\textbf{Voxelization.} 
Let $\mathcal{C}$ denotes the 3D overall space with sizes of $(D,W,H)$ along the axes of $x$, $y$ and $z$.
As illustrated in Fig.\ref{fig:framework}, 
we first divide the space into equal-sized cells ${c}^i$ with sizes of $(D/n_D, W/n_W,H/n_H)$.
Then, the points $\mathbf{p}_i$'s in the point cloud $P$ are dispensed to the respective cells and we refer to these cells as voxels in the following.  
Due to the sparse nature of LiDAR data, most voxels are only allocated $2$-$3$ points with the voxel size of ($10$ cm, $10$ cm, $20$ cm), which achieves a good balance between representation precision and resource cost.

\textbf{Feature Encoding.}
Each LiDAR point $\textbf{p}^i\in S$ is a vector of $[x^i,y^i,z^i,n_x^i, n_y^i, n_z^i, r^i]^T$, where $(x^i,y^i,z^i)$ is the $i$-th point's coordinate, $(n_x^i, n_y^i, n_z^i)$ is its normal vector (which can be obtained  via cross products over four neighbors similar to \cite{li2019net,chounsupervised}) and $r_i$ stands for the reflectance value. 
For computational efficiency, we compute the arithmetic average $\bar{\mathbf{p}}^j=1/|c^j|\sum_{\mathbf{p}^i\in c^j} \mathbf{p}^i$
for each voxel as its representation and obtain down-sampled point clouds $P_t$ and $P_{t-1}$, which will be used as the initial point cloud's representation.
Thereafter, we extract the high level features from the point cloud through stacked 3D submanifold convolutions~\cite{graham2017submanifold} illustrated in Fig.~\ref{fig:framework}.
The 3D submanifold convolution, combined of valid sparse convolution and sparse convolution, has the advantages of accurately capturing 3D local geometric patterns of the input point cloud while ignoring the empty voxels to accelerate the encoding process. 
The 3D convolution maintains the geometric structures and spatial topology during down-sampling/up-sampling, which is quite challenging for the 2D convolution as shown in Fig.~\ref{fig:3d_conv}.
Finally, we obtain the voxel feature volumes
$V_{t-1}$ and $V_{t}$ as an intermediate high-level representation for the point clouds $P_{t-1}$ and $P_{t}$ respectively.

\subsection{Odometry Regression}\label{sec:odom_regress}
\vspace{-1ex}

After encoding the voxel features 
$V_{t-1}$ and $V_{t}$ 
from the input point clouds, we design a network to predict the ego-motion from these features. 
To reduce the memory cost and improve the computational efficiency, we follow \cite{zhou2018voxelnet} to reshape the 3D features volume of the two timestamps ${t-1}$ and $t$ to the 2D bird-view feature maps respectively and channel-wisely concatenate the features of the two timestamps before feeding them into the ego-motion regression module.
Thanks to the powerful encoding capability of 3D CNNs, the features can successfully encode the agent's motion 
in both the ground plane and the vertical direction.

The ego-motion regression module is constituted by several ResNet~\cite{he2016deep} blocks followed by fully connected layers, which estimates the ego-motions $(\mathbf{R}^{pred}, \mathbf{t}^{pred})$ as illustrated in Fig.~\ref{fig:framework}.
In our implementation, we let the network predict the 
quaternion vector,
a more compact representation for rotation, which is more suitable as the regression target.
Since the quaternion representation and the matrix representation are equivalent, we uniformly refer to the rotation prediction as $\mathbf{R}^{pred}$  in the following for simplicity.

\textbf{Spherical Reprojection Loss.} 
Let $O_{t-1}$ and $O_{t}$ be the agent coordinate systems at two consecutive timestamps and $\mathbf{x}$ denote an arbitrary static point in the 3D scene. As illustrated in Fig~\ref{fig:framework}a. The observations of this point in systems $O_{t-1}$ and $O_t$
are denoted as $\mathbf{x}_{t-1}$ and $\mathbf{x}_{t}$ respectively. 
Ideally, we can obtain $\mathbf{x}_t$ as
\begin{equation}\label{eq:sph_loss_transform}
    \mathbf{x}_t = \mathbf{R}^{gt}\mathbf{x}_{t-1}+\mathbf{t}^{gt},
\end{equation}
where $\mathbf{R}^{gt}$ and $\mathbf{t}^{gt}$ are the ground-truth ego-motion, which however is not available in our setup.
To achieve self-supervised learning of ego-motion prediction, our network leverages the geometric consistency among frames: to first transform the point cloud $P_{t-1}$ with the current prediction ($ \mathbf{R}^{pred},  \mathbf{t}^{pred}$) obtaining $P'_t$, \ie, transform each point $\mathbf{x}_{t-1}\in P_{t-1}$ as  $\mathbf{x}'_t = \mathbf{R}^{pred}\mathbf{x}_{t-1}+\mathbf{t}^{pred}$, and then minimize the distance between $P'_t$ and $P_t$
to push the prediction closer to $(\mathbf{R}^{gt}, \mathbf{t}^{gt})$.
The nearest-neighbor Euclidean distance is a common choice to point-wisely measure the deviation between the two point clouds. However, we find that the Euclidean distance is not an optimal choice to directly measure the discrepancy between the two 3D point clouds, because of the high sparsity level and less measurement accuracy in the distance caused by the sparsity nature of the LiDAR point clouds.
Due to the noise and the erroneous correspondence identifying caused by the increasing sparsity level, the nearest-neighbor Euclidean distances between the two point clouds are much larger in distant regions than those of the nearby regions, which are illustrated in Fig.~\ref{fig:distribution}.
As a result, the distant unreliable points might contribute more in the loss function with Euclidean distance measurement, which is not desirable.
Inspired by the reprojection error~\cite{hartley2003multiple} widely used in visual odometry, we propose 
the spherical reprojection loss to make the loss focusing more on the nearby reliable points.
Specifically, we adopt the pair-wise angular distance to measure the distance between the 
associated points with nearest-neighbor searching from
the point clouds $P_t$ and $P'_t$, where the angular distance is defined as the angle between origin-to-point rays of associated correspondence points in $P_t$ and $P_{t-1}$. 
As shown in Fig.~\ref{fig:framework}a, the angular distance between the nearest-neighboring points $\mathbf{x}_t$ and $\mathbf{x}'_t$ is defined as the angle $\theta$ between the rays $O_tG_t$ and $O_tG'_t$, where $G_t$ and $G'_t$ are the projection of nearest-neighboring points on a unit sphere centered at $O_t$.
The minimization of this angle error is equivalent to minimize the geodesic distance on the sphere, which is similar to the minimization of reprojection error.
For numeric stability, 
we minimize the $-\cos(\theta)$, a monotonically increasing function of $\theta$ in $[0,\pi]$. The spherical reprojection loss is hence expressed as   
\begin{equation}\label{eq:sph_loss}
	L_{sr} = -\frac{1}{|P_t|}\sum_{\mathbf{x}_t^{i}\in P_{t}, j=\mathcal{M}(i)}\left(\frac{\mathbf{x}^{i}_t\cdot\mathbf{x}'^{j}_t}{ ||\mathbf{x}^{i}_t||_2||\mathbf{x}'^{j}_t||_2}\right),
\end{equation}
where $\mathcal{M}(\cdot)$ denotes the nearest-neighbor 
identifying process
between the points in $P_t$ and $P'_{t}$.  As illustrated  by the green curve in Fig.~\ref{fig:distribution}, the nearby errors contribute more to the spherical reprojection loss, which makes the loss focus more on the nearby reliable regions.

\textbf{Transformation Residual Loss.}
To stabilize and speed up the convergence, we further incorporate the classic ICP algorithm~\cite{chen1992object,low2004linear} into the loss function to directly guide the ego-motion estimation learning. 
As illustrate in Fig.~\ref{fig:framework}b, we first align the point cloud $P_{t-1}$ to $P'_{t}$ with the current ego-motion prediction as in the previous section, 
and then calculate transformation residual $(\delta\mathbf{R},\delta \mathbf{t})$ from $P'_t$ to $P_t$ with ICP iteration.
The ICP iteratively finds the nearest-neighbor correspondences $(\mathbf{x}^i_{t},\mathbf{x}'^j_t)$'s and minimize the point-to-plane distances $\mathcal{D}(\mathbf{x}^i_{t},\mathbf{x}'^j_t )$ between them. 
By accumulating the computed transformation residual to the current prediction ($\mathbf{R}^{pred}$,$\mathbf{t}^{pred}$), we can obtain a more accurate ego-motion: 
\begin{equation}\label{eq:rectified_ego}
	{\mathbf{t}^*}=\delta \mathbf{R}\mathbf{t}^{pred}+\delta\mathbf{t},~~~\mathbf{R}^*=\delta \mathbf{R}\mathbf{R}^{pred}
\end{equation}
To improve the prediction accuracy, we design our transformation residual loss as \footnote{In practical implementation, $L_{tr} = u_{\alpha}(||( \mathbf{t}^* - \mathbf{t}^{pred}||^2_2) + u_{\beta}(|| \mathbf{R}^*-\mathbf{R}^{pred} ||^2_F)$. $\mathbf{t}^*$ and $\mathbf{R}^*$ are regarded as targets and the back-propagating gradients are stopped at $\mathbf{t}^*$ and $\mathbf{R}^*$.  } 
\begin{equation}\label{eq:res_improve_loss}
  L_{tr} = u_{\alpha}(||(\delta\mathbf{R}-\mathbf{I})\mathbf{t}^{pred} +\delta\mathbf{t}||^2_2) + u_{\beta}(||\delta \mathbf{R}-\mathbf{I}||^2_F),
\end{equation}
to push $\mathbf{t}^{pred}\rightarrow{\mathbf{t}^*}$ and $\mathbf{R}^{pred}\rightarrow{\mathbf{R}^*}$,
where $u_\diamond(\cdot)$ is a uncertainty-aware loss: $u_\diamond(l)=e^{-\diamond}l+\diamond$, to model the homoscedastic noise during training~\cite{kendall2017geometric}, where $\diamond$ is a learnable parameter.
Ideally, the transformation residual ($\delta \mathbf{R}, \delta \mathbf{t}$) should be close to an identity transformation ($\mathbf{I}$, $\mathbf{0}$) after the training converges.

\textbf{Deep Flow Supervision Loss.}
All the above losses are used to directly supervise the final ego-motion estimation.
To further enhance the point-wise feature representations, we incorporate the 3D scene flow prediction as an auxiliary task in a unsupervised manner.
The flow prediction highly relies on the local topology patterns, which could makes the features encode more geometric information.
As shown in Fig.~\ref{fig:framework}, 
we add several scene flow prediction branches (with several convolution layers) at the different depths of the ego-motion regression encoder to predict scene flow ${F}^{pred}$ from the voxel features (each feature vector 
can be mapped to a 3D voxel location $\mathbf{x}^i_t$ with our voxel-based representation).
The flow of a point $\mathbf{x}_t^i$ is defined as its coordinate difference between two timestamps: ${F}_h(\mathbf{x}_t^i)=\mathbf{x}_{t}^i-\mathbf{x}_{t-1}^i=(\mathbf{I} - \mathbf{R}^{gt})^\mathrm{T}\mathbf{x}_{t}^i+(\mathbf{R}^{gt})^\mathrm{T}\mathbf{t}^{gt}$. 
Since the ego-motion ground-truth is not available, we approximate the scene flow targets with the rectified ego-motion prediction with  the transformation residuals obtained by the ICP iteration according to Eq.~\eqref{eq:rectified_ego}, 
and then calculate the scene flow of the voxels:
 $h\in\mathcal{H}$: ${F}^*_h(\mathbf{x}_{t}^{i})=(\mathbf{I} - \mathbf{R}^{*})^\mathrm{T}\mathbf{x}_{t}+(\mathbf{R}^{*})^\mathrm{T}\mathbf{t}^{*}$,
where $\mathbf{x}_t^{i}$ denotes the $i$-th voxel center's coordinate in encoder layer $h$.
Finally, we take this approximated flow as the target to supervise the flow prediction at different encoder depths:
\begin{equation}\label{eq:flow_loss}
  L_{fs} = 
  \sum_{h, i} w_h\cdot u_{\alpha}(||{F}^{pred}_h(\mathbf{x}_t^{i} ) -{F}^*_h(\mathbf{x}_t^{i} ) ||^2_2),
\end{equation}
where $w_h$ is the weight for layer $h$, and $u_{\alpha}(\cdot)$ is a uncertainty-aware loss mentioned before.

\subsection{Correspondence Confidence Estimation }
In practical scenarios, the nearest-neighbor-based correspondence mapping $\mathcal{M}(\cdot)$ used in the Eq.~\eqref{eq:sph_loss} and Eq.~\eqref{eq:res_improve_loss} to associate corresponding points in two timestamps are not always accurate, because of the existence of moving objects, noise, and measurement errors.
To alleviate the adverse effects from these inaccurate correspondences, 
we design a 3D decoder sub-network following the geometric feature encoder to estimate the reliability for the points, which is implemented
as 3D transposed convolution layers followed by sigmoid functions squeezing the output range to $[0,1]$.
As illustrated in Fig.~\ref{fig:framework}, 
the confidence estimation decoder estimates the point-wise confidence $M'_t=\{{m}_t'^i| i=1,
\ldots,N\}$ (trivially obtained from $M_t$) and $M_{t}=\{{m}_t^j| j=1,\ldots,N\}$ for point sets $P_{t}'$ and $P_t$ respectively.
We takes pair-wise product
$\{\mathrm{M}^{ij}={m}_t'^i{m}_t^j|i=1,\ldots,N,j=\mathcal{M}(i) \}$ 
to estimate the reliability of each matched correspondence pair 
($\mathbf{x}_t^i, \mathbf{x}_t'^j$) needed by Eq.~\ref{eq:sph_loss} and the ICP optimization for Eq.~\ref{eq:res_improve_loss}. 
The confidence factors can be straightforwardly incorporated into Eq.~\eqref{eq:sph_loss}: 
\begin{equation}\label{eq:mask_sph_loss}
	L_{sr} = -\frac{1}{|P_t|}\sum_{\mathbf{x}_t^{i}\in P_{t}, j=\mathcal{M}(i)}\frac{\mathrm{M}^{ij}\cdot \mathbf{x}^{i}_t\cdot\mathbf{x}'^{j}_t}{ ||\mathbf{x}^{i}_t||_2||\mathbf{x}'^{j}_t||_2}. 
\end{equation}
For the transformation residual loss (Eq.~\eqref{eq:res_improve_loss}), we modify the original ICP optimization term as 
\begin{equation}\label{eq:icp_term}
    E=\frac{1}{|P_t|} \sum_{\mathbf{x}_t^{i}\in P_{t}, j=\mathcal{M}(i)} (\mathrm{M}^{ij}/\max_i(\mathrm{M}^{ij})+\epsilon ) \cdot\mathcal{D}(\mathbf{x}^i_{t},\mathbf{x}'^j_t ),
\end{equation}
where $\mathcal{D}(\cdot,\cdot)$ denotes the distance function in ICP, $(\mathbf{x}^i_{t},\mathbf{x}'^j_t)$ are the identified correspondences between $P_t$ and $P'_t$ with the nearest-neighbor mapping $\mathcal{M}(\cdot)$, and the constant addend $\epsilon$ (empirically set to $0.1$ in our experiments) is to avoid the extreme imbalance of weights in early training phases. 
The classic ICP assumes that all correspondence pairs can be matched perfectly by optimizing the transformation ($\delta\mathbf{R}$,$\delta\mathbf{t}$) in Fig.~\ref{fig:framework}b, which is often not true especially when moving objects/noises exist.  The correspondence confidence factors successfully handle this dilemma to lower the weights on the unreliable correspondences during optimization.

Since there is no ground-truth for the confidence prediction, we use the self-supervised range alignment error to guide this confidence estimation similar to \cite{zhou2017unsupervised}:
\begin{equation}\label{eq:mask_range_loss}
	L_{ra} =  \frac{1}{|P_t|}\sum_{\mathbf{x}_t^{i}\in P_{t}, j=\mathcal{M}(i)}\mathrm{M}^{ij}(r(\mathbf{x}^{i}_t)-r(\mathbf{x}'^{j}_t))^2-\gamma\log(\mathrm{M}^{ij}),  
\end{equation}
where $r(\cdot)=||\cdot||_2$  calculates the range value and the regularization term $-\log(\mathrm{M}^{ij})$ with weight $\gamma$  avoids the all-zero trivial prediction.

In summary, our overall loss function is finally expressed as 
\begin{equation}\label{eq:final_loss}
  L = w_1L_{sr} + w_2L_{ra} + w_3L_{tr}+ w_4L_{fs},
\end{equation}
where $w_1$-$w_4$ are the weights of different losses.


\section{Experimental Results}
\subsection{Benchmark Dataset and Evaluation Metrics}
\textbf{KITTI Odometry Dataset.}
KITTI Odometry dataset\cite{geiger2012we} consists of 22 LiDAR sequences with corresponding color/gray images. It provides ground-truth poses for sequences 00-10 obtained from IMU/GPS fusion algorithms. 
The remaining sequences are for benchmark testing and do not provide ground-truth poses. This dataset covers different types of road environments (including country roads, urban areas, highways \etc), and contains pedestrains, cyclists and different types of vehicles. The speed of acquisition vehicles varies in different areas ranging from 0 km/h to 90 km/h.
\\
\textbf{Apollo-SouthBay Dataset.}
Apollo-SouthBay Dataset~\cite{lu2019l3} collected in the San Francisco Bay area, United States, covers various scenarios including residential area, urban downtown areas and highways. It also provides the ground-truth poses and training/testing splits for the first five scenarios which are adopted in our experiments. The apollo sequence is longer than the KITTI's and the scenarios are more complicated, which is suitable for testing the generality of our method.   \\
We adopt the official evaluation metrics provide by the KITTI benchmark~\cite{geiger2012we} to measure the translational and rotational drift on length of 100m-800m  in our experiments.
The implementation details of our method can be found in supplementary materials.

\subsection{Comparison with State-of-the-arts}
\begin{table}[t] 
    \caption{Comparison with the state of the arts on KITTI odometry dataset. We also list 
    the supervised methods and visual odometry methods here for reference. 
    }
    \label{tab:eval_kitti}
      \centering
      \tiny
      \resizebox{0.95\linewidth}{!}{
      \begin{tabular}{r|cccccccccccccc}
        \hline
        &\multirow{2}{*}{Seq.} & \multicolumn{2}{|c|}{Training Seq.}
        &\multicolumn{2}{c|}{7}&
        \multicolumn{2}{c|}{8}
        &\multicolumn{2}{c|}{9}
        &\multicolumn{2}{c|}{10} 
        &\multicolumn{2}{c}{Testing Avg.} \\
        \cline{3-14}
        &&\multicolumn{1}{|c}{$t_{rel}$}&$r_{rel}$ 
        &$t_{rel}$&$r_{rel}$ 
        &$t_{rel}$&$r_{rel}$
        &$t_{rel}$&$r_{rel}$
        &$t_{rel}$&$r_{rel}$
        &$t_{rel}$&$r_{rel}$ \\
        \hline\hline
        \multirow{7}{*}{\rotatebox{90}{Classic} } &ICP-po2po &6.45&3.16&5.17 &3.35 & 10.04&4.93&6.93&2.89&8.91&4.47 &7.76&3.98\\
        &ICP-po2pl &3.76&1.79&1.55 &1.42 & 4.42&2.14&3.95&1.71&6.13&2.60&4.01&1.97 \\
        &{GICP~\cite{segal2009generalized}} &1.87&0.76&0.64 &0.45 & 1.58&0.75&1.97&0.77&1.31&0.62&1.38&0.65\\
        &{NDT-P2D~\cite{stoyanov2012fast}} &34.2&5.73&7.51&3.07&13.6&4.62&33.7&7.06&20.5&3.06&18.8&4.45\\
        &{CLS~\cite{velas2016collar}} &2.30&0.93&1.04 &0.73 & 2.14&1.05&1.95&0.92&3.46&1.28&2.15&1.00 \\
        &{LOAM~(w/o mapping)} &3.90&1.53&2.98 &1.55 & 4.89&2.04&6.04&1.79&3.65&1.55&4.39&1.73 \\
        &{LOAM~(w/ mapping)~\cite{zhang2014loam}} &1.12&0.50&0.69 &0.50 & 1.18&0.44&1.20&0.48&1.51&0.57&1.15&0.50 \\
        \hline
        \multirow{3}{*}{\rotatebox{90}{Sup.} }&{Velas~\etal\cite{velas2018cnn}}  &2.90&-&1.77 &- & 2.89& - &4.94& - &3.27&-&3.22&- \\
        &{LO-Net~(w/o mapping)~\cite{li2019net}}  &1.05&0.66&1.70 &0.89 & 2.12&{0.77}&1.37&0.58&1.80&0.93&1.75&0.79 \\
        &{LO-Net~(w/ mapping)~\cite{li2019net}}  &0.71&0.45&0.56 &0.45 & {1.08}&0.43&0.77&0.38&{0.92}&0.41&{0.83}&0.42 \\
        \hline
        \multirow{6}{*}{\rotatebox{90}{Unsup.} }&Zhou~\etal*~\cite{zhou2017unsupervised}  &30.8&4.63&21.3 &6.65 & 21.9&2.91&18.8&3.21& 14.3&3.30&19.1&4.02 \\ 
        &UnDeepVO*~\cite{li2018undeepvo}  &4.80&2.69&3.15 &2.48 & 4.08&1.79&7.01&3.61& 10.6&4.65&6.22&3.13 \\ 
        &{Cho~\etal~\cite{chounsupervised}}  &3.68&0.87&-&- & -& - &4.87&1.95& 5.02&1.83&4.95&1.89 \\ 
        \cline{2-14}
        &Ours &2.50&1.11&3.09&1.81&3.16 & 1.14 &3.01&1.14& 3.48&1.11&3.19&1.30 \\ 
        &Ours~(more data) &2.31&1.06&2.51&1.15&2.65 & 1.00 &2.86&1.17& 3.22&1.26&2.81&1.15 \\ 
        &Ours~(more data, w/mapping) &\textbf{0.70}&\textbf{0.37}&\textbf{0.34}&\textbf{0.21}&{1.18} & \textbf{0.35} &\textbf{0.83}&\textbf{0.34}& \textbf{1.22}&\textbf{0.40}&\textbf{0.89}&\textbf{0.32} \\ 
        \hline 
      \end{tabular}
      }
  
      {
        \centering
        ~~~~$t_{rel}, r_{rel}$: Average translational RMSE (\%) and rotational RMSE ($^\circ$/100m) on length of 100m-800m~\cite{geiger2012we}. *: Visual odometry methods \\
  
      \par}
      \vspace{-4ex}
  \end{table}

\begin{figure}
    \centering
    \begin{subfigure}[t]{0.473\textwidth}
        \includegraphics[width=\linewidth, trim=1.9cm 1.5cm 19cm 12.5cm, clip]{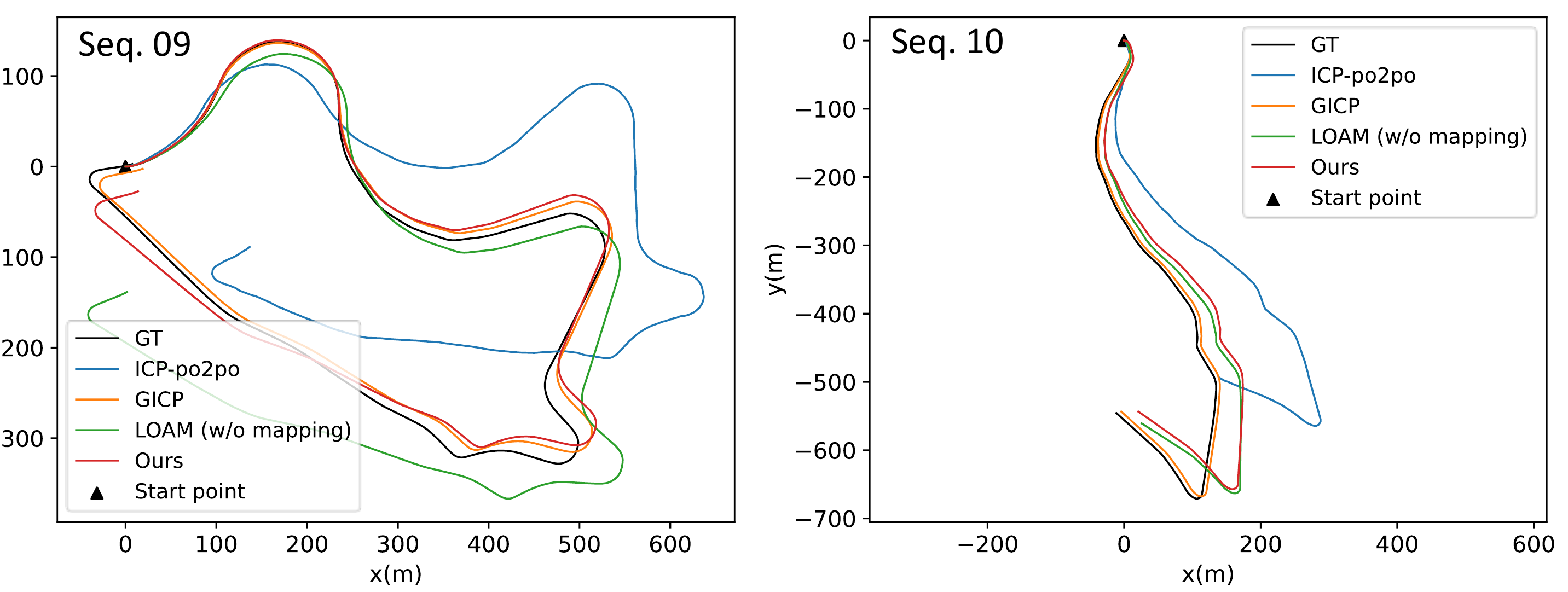}
    \caption{}\label{fig:traj_plot}
    \end{subfigure}
\begin{subfigure}[t]{0.24\textwidth}
    \centering
    \includegraphics[width=1\linewidth, trim=0cm 13cm 35cm 1cm, clip]{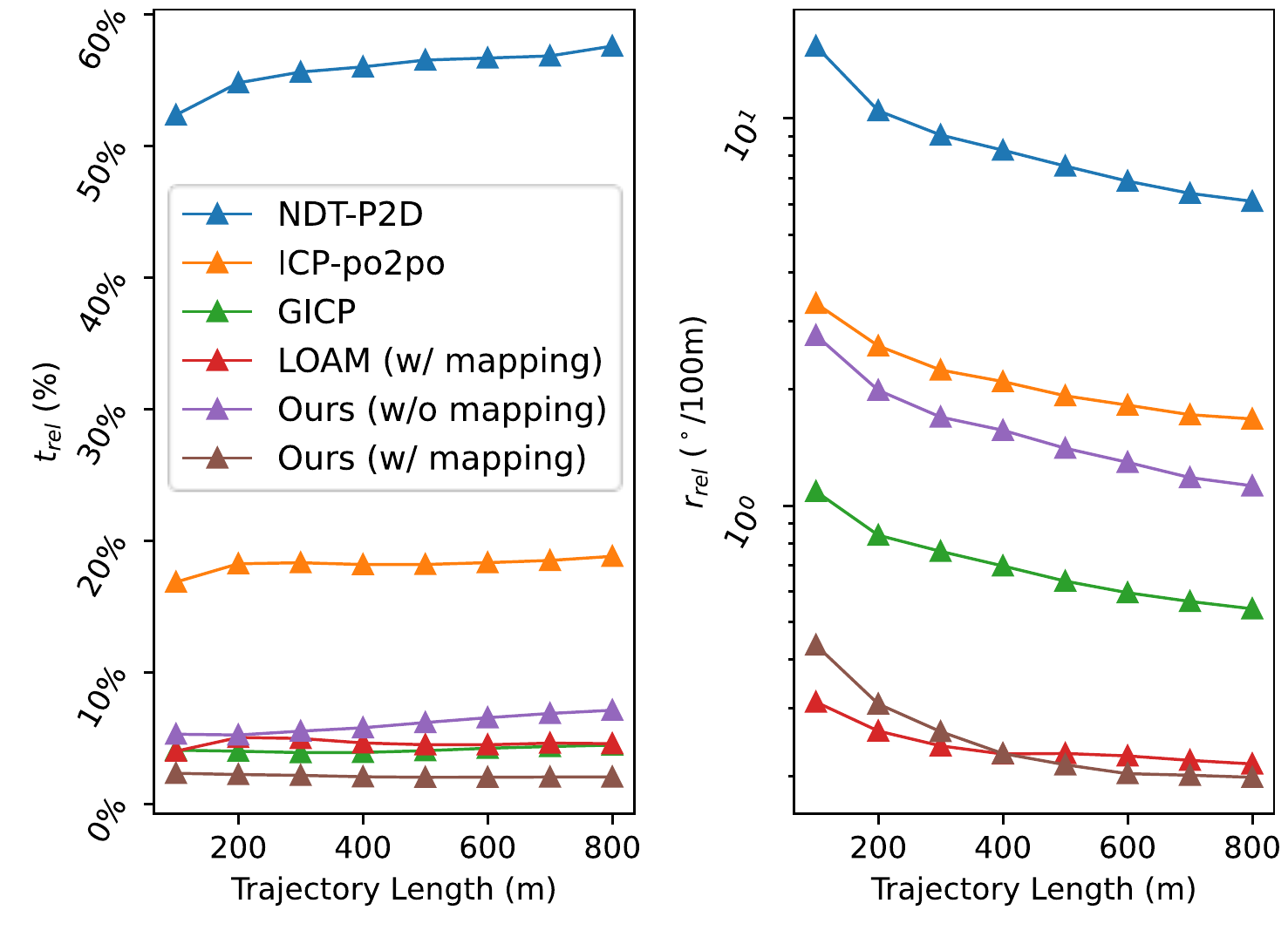}
    \caption{}\label{fig:seg_err}
\end{subfigure}
\begin{subfigure}[t]{0.24\textwidth}
    \centering
    \includegraphics[width=1\linewidth, trim=0cm 13cm 35cm 1cm, clip]{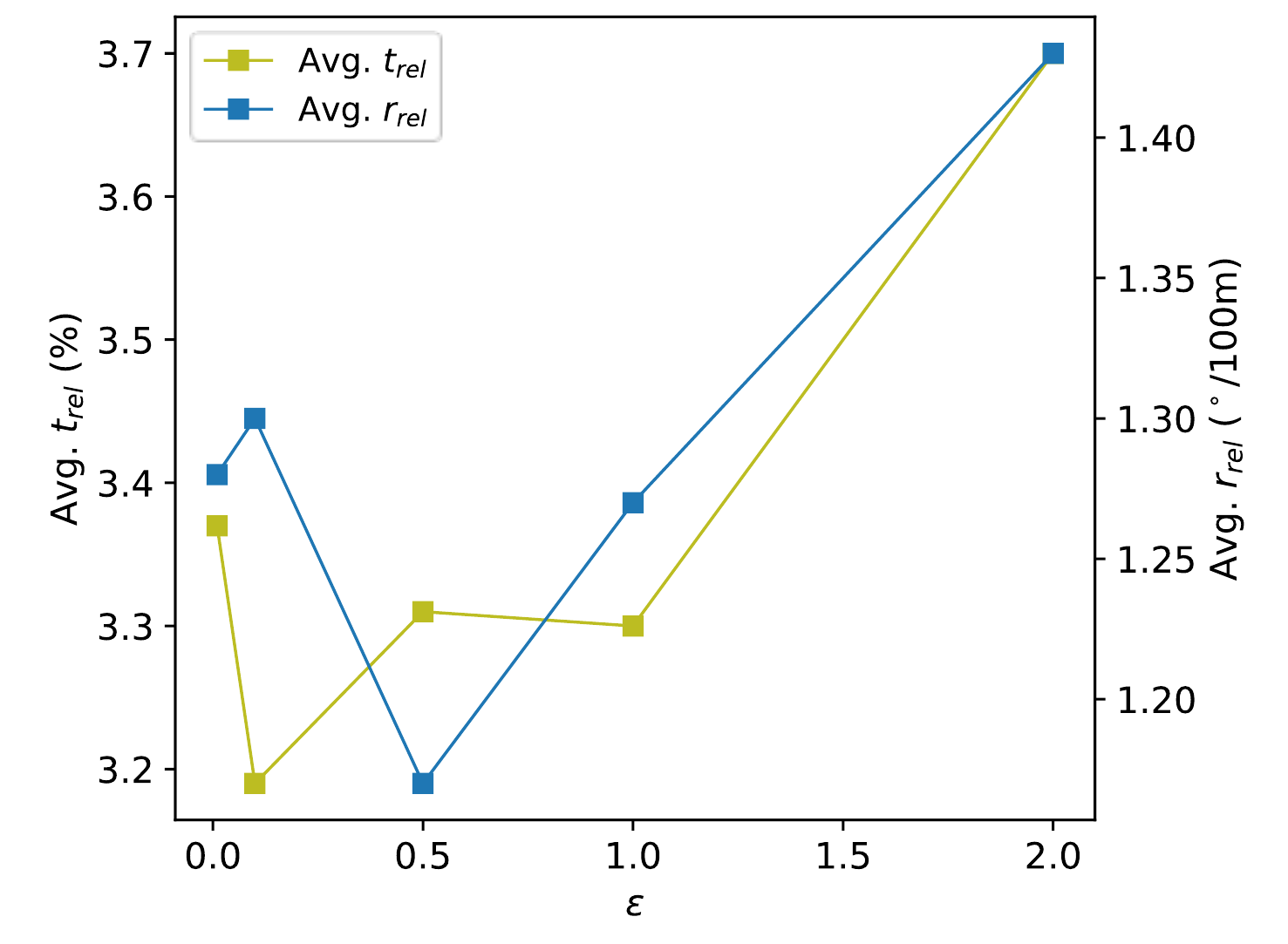}
    \caption{}\label{fig:eps}
\end{subfigure}
    \caption{(a)~Trajectory plots comparison among some two-frame methods on KITTI. (b)~The average translation and rotation errors with respect to trajectory length intervals  for different methods on Apollo.
    (c)~Performance variation of our method on KITTI test set with different $\epsilon$ values. }
\vspace{-3ex}
\end{figure}

\textbf{Evaluation on KITTI Dataset.} We compare our method with other competitive methods on the sequences 00-10 of the KITTI odometry dataset. These compared methods adopted different splitting strategies for training and testing 
\footnote{\cite{velas2018cnn} takes the sequences 00-07/08-10 as the train/test split, \cite{li2019net} takes 00-06/07-10 as the train/test sets, while the others CNN-based methods takes 00-08/09-10 for training/testing}.
\textit{Since most of their code are not available, for fair comparison, we choose the splitting strategy with minimal training data (\ie~00-06/07-10 for training/testing) to evaluate our method against them}. 
Table \ref{tab:eval_kitti} shows the evaluation results where we exclude the results of Seq. 01 following other unsupervised methods~\cite{chounsupervised}. The Seq.~01 on highway is in a very open space with few structures to infer the agent's motions, where most of the unsupervised method and classic methods fail.
Compared with other state-of-the-art unsupervised methods, our method achieves the best performance (denoted as `Ours' in Table \ref{tab:eval_kitti}) even with the least amount of data for training.
We also try to add the more unlabeled (Seq. 11-21) into the training set (denoted as `Ours (more data)'), and our performances are further improved and even surpasses those of some supervised methods, which shows our unsupervised method's scalability to be benefited from more training data. 
We also compare our method with classic methods~\cite{arun1987least,stoyanov2012fast,segal2009generalized,velas2016collar}. 
Our method achieves significant better performance than most two-frame methods widely used, \ie, point-to-point ICP \cite{arun1987least}, point-to-plane ICP~\cite{chen1992object} and NDT-P2D \cite{stoyanov2012fast}, and reasonably 
inferior performance than the time-consuming (only $\sim0.5$Hz) GICP~\cite{segal2009generalized} and CLS~\cite{velas2016collar} which iteratively refine the pose transformation with computational heavily correspondence association modeling. (Specific runtime comparison can be found in the supplementary materials.) {Fig.~\ref{fig:traj_plot} plots example trajectories in Seq.~09-10 of different two-frame methods for visualization.}  Although we mainly focus on the two-frame ego-motion estimation in this paper, we also incorporate the mapping module from LOAM~\cite{zhang2014loam} into our framework to test the odometry performance with the additional backend multi-frame refinement. The results (denoted as `Ours~(more data, w/ mapping)') in Table~\ref{tab:eval_kitti} demonstrate that our method can be successfully coupled with multi-frame refinement widely used in SLAM systems. We achieves better performance than LOAM, a competitive classic LiDAR-based SLAM system, and show comparable performance with state-of-the-art supervised odometry method LO-net \cite{li2019net} with mapping refinement.

\textbf{Evaluation on Apollo-SouthBay Dataset.}We also evaluate our model on more challenging Apollo-SouthBay dataset to further demonstrate our generality. As shown in Table~\ref{tab:eval_apollo} and Fig.~\ref{fig:seg_err}, our two-frame-based network prediction (denoted as `Ours') consistently outperform most of the classic two-frame-based methods \cite{arun1987least,chen1992object,stoyanov2012fast} and achieves comparable translational accuracy to the multi-frame-based LOAM~\cite{zhang2014loam}. We also prove that our method with mapping refinement can outperform other methods. The Apollo dataset is closer to actual autonomous driving scenarios with more moving objects. By training on large-scale data and correspondence confidence estimation, our method shows consistent robustness.

  \begin{table}[t] 
    \caption{The evaluation results on Apollo-SouthBay test set.}
    \label{tab:eval_apollo}
    \begin{center}
      \tiny
      \resizebox{0.85\linewidth}{!}{
      \begin{tabular}{r|cccccccccccccccccccccccc}
        \hline
        & \multicolumn{2}{c|}{ICP-po2po} 
        & \multicolumn{2}{c|}{ICP-po2pl} 
        & \multicolumn{2}{c|}{GICP~\cite{segal2009generalized}} 
        & \multicolumn{2}{c|}{NDT-P2D~\cite{stoyanov2012fast}} 
        & \multicolumn{2}{c|}{LOAM(w/ mapping)~\cite{zhang2014loam}} 
        & \multicolumn{2}{c|}{Ours}
        & \multicolumn{2}{c}{Ours (w/ mapping)}\\
        \cline{2-15}
        &$t_{rel}$&$r_{rel}$ 
        &$t_{rel}$&$r_{rel}$ 
        &$t_{rel}$&$r_{rel}$ 
        &$t_{rel}$&$r_{rel}$ 
        &$t_{rel}$&$r_{rel}$ 
        &$t_{rel}$&$r_{rel}$ 
        &$t_{rel}$&$r_{rel}$ \\
        \hline
        Avg.&22.8 &2.35  	
         & 7.75&1.20 	
         & 4.55	& 0.76 
        &57.2 & 9.40
        &5.93 &0.26 	
        &	6.42& 1.65
        &\textbf{2.25}&\textbf{0.25}\\
        \hline 
      \end{tabular}
      }
  
      \vspace{-4ex}
    \end{center}
  \end{table}

\subsection{Ablation Study}
\begin{table}[t] 
    \caption{Comparison among different ablation variants.}
    \label{tab:ablation}
    \begin{center}
      \resizebox{0.99\linewidth}{!}{
      \begin{tabular}{c*{12}{|P{1.6cm}} }
        \hline
         & \multicolumn{2}{c|}{$L_{sr}$}
         & \multicolumn{2}{c|}{$L_{sr}, L_{ra}$}
         & \multicolumn{2}{c|}{$L_{sr}, L_{ra},L_{tr}$} 
         & \multicolumn{2}{c|}{$L_{sr}, L_{ra},L_{tr}, L_{fs}, w/o~conf.$} 
         & \multicolumn{2}{c|}{$L_{eu}, L_{ra},L_{tr}, L_{fs} $} 
         & \multicolumn{2}{c}{$L_{sr}, L_{ra},L_{tr}, L_{fs}$} \\
        \hline
        &train&test
        &train&test
        &train&test
        &train&test
        &train&test
        &train&test\\
        \hline
        Avg. $t_{rel}$ 
        &12.4&19.1
        &4.63&7.19
        &2.59&4.84 %
        &3.24&3.59 %
        &2.88&3.50
        &\textbf{2.50}&\textbf{3.19}\\ %
        Avg. $r_{rel}$ 
        &4.72&6.54
        &1.90&2.65
        &1.20&1.96 %
        &1.82&2.04 %
        &1.41&1.27
        &\textbf{1.11}&\textbf{1.30}\\%
        \hline 
      \end{tabular}
      
      }
      \vspace{-4ex}
    \end{center}
  \end{table} 
To verify the effectiveness of each proposed module and unsupervised loss functions, we conduct a throughout ablation study on the KITTI dataset as shown in Table~\ref{tab:ablation}. We test different combinations of loss functions incrementally and the proposed full loss to show that it has optimal performance compared with losses with fewer terms. 
We also try to remove our confidence mechanism from our model (denoted as `w/o conf.'), and we can see evident performance drop due to  unreliable correspondences caused by noises, dynamic objects and varying pointcloud densities. 
We visualize some estimated confidence examples in Fig.~\ref{fig:confidence}, where our network successfully lowers the confidences of points on dynamic objects, \eg~vehicles, cyclists \etc~and has higher confidence on the static poles and vertical surfaces.
Moreover, we substitute the Euclidean loss $L_{eu}$ for our spherical reprojection loss $L_{sr}$ and find obvious performance drop.
Besides, in Fig.~\ref{fig:eps}, we analyse the the performance variations with different $\epsilon$ values in Eq.~\ref{eq:icp_term}  and find that large $\epsilon$ values lead to inferior performances, as the large $\epsilon$ weakens the effect of estimated confidence weights.

\begin{minipage}{\textwidth}
  \centering 
  \begin{minipage}[]{0.49\textwidth}
    \centering
    \includegraphics[width=0.65\linewidth, trim=2cm 9.1cm 28cm 1cm, clip]{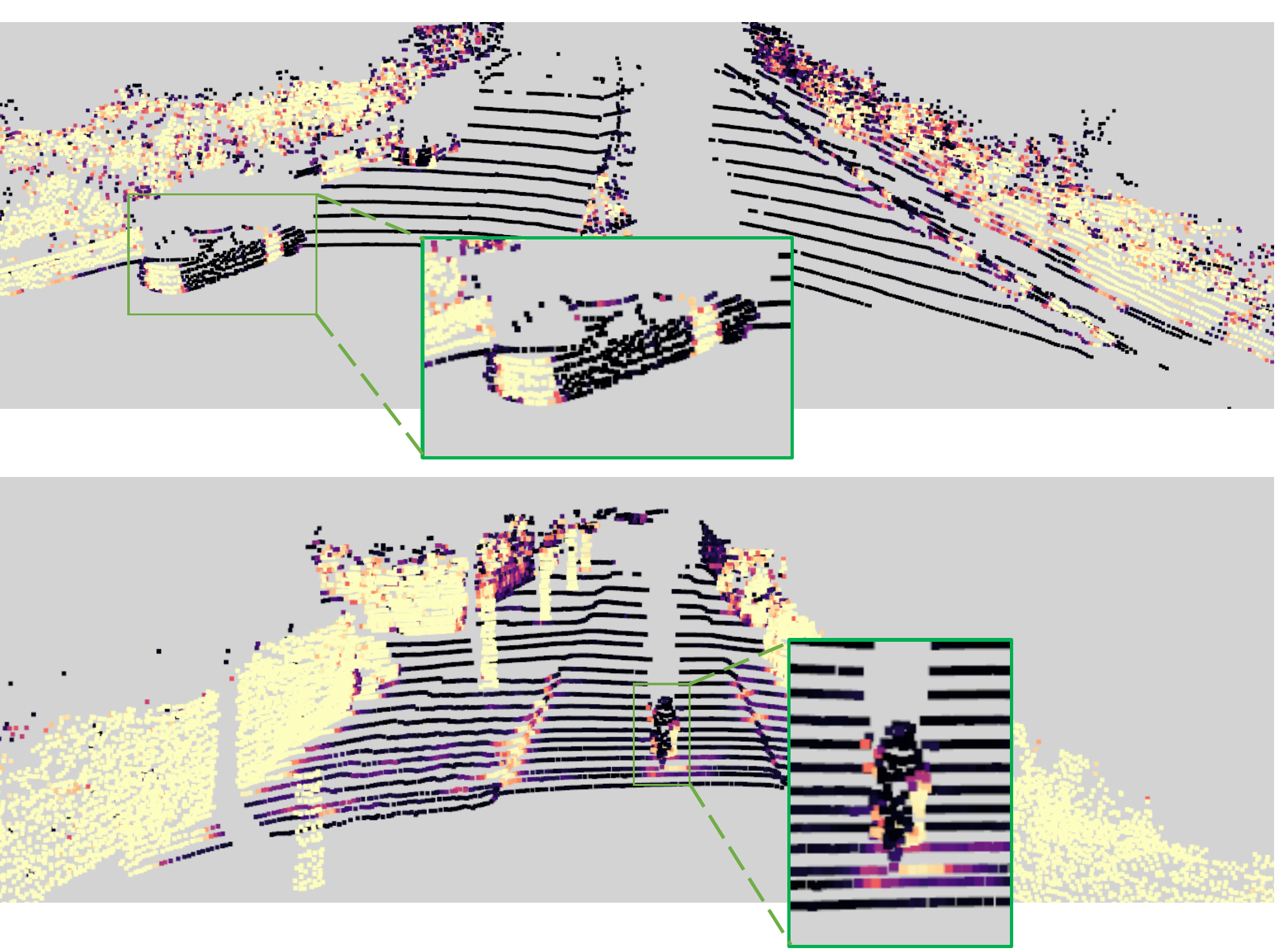}
    \captionof{figure}{Our confidence prediction visualization. The brighter denotes higher confidendence. }\label{fig:confidence}
  \end{minipage}
  \begin{minipage}[]{0.5\textwidth}
    \centering
    \parbox{1\linewidth}{
    \centering
     \captionof{table}{The runtime of our submodules and the comparison with other methods.} \label{tab:runtime}
    \resizebox{0.6\linewidth}{!}{
     \begin{tabular}{c*{1}{|c}}
    \multicolumn{2}{c}{\textit{\textbf{Our submodule runtime}}} \\
    \hline
      Module & Time (ms) \\
      \hline
      Voxelization & 0.7\\
      Voxel Feature Extraction & 65.8\\
      Odometry Regression & 21.9 \\
      \hline
      Total & 88.4\\
      \hline
    \end{tabular}
    }
    \vspace{1ex}
    \resizebox{0.6\linewidth}{!}{
    \begin{tabular}{c|*{2}{c}}
    \multicolumn{2}{c}{\textit{\textbf{Runtime comparison with other methods}}} \\
      \hline
      Methods & Time (ms) \\
      \hline 
      ICP-po2po~\cite{arun1987least} & 261\\
      ICP-po2pl~\cite{chen1992object} & 1250\\
      GICP~\cite{segal2009generalized} & 1781\\
      NDT-P2D~\cite{stoyanov2012fast} & 1723\\
      CLS~\cite{velas2016collar}& 19843 \\
      Ours&88.4 \\
      \hline
    \end{tabular}
    }
    }
   
    \end{minipage}
  \end{minipage}

\vspace{-3ex}
\subsection{Runtime Analysis}\label{sec:runtime}
\vspace{-1ex}
We further analyse the running time of our framework on a machine with a Xeon(R) E5-2697A v4 CPU and a NVIDIA Tesla V100 GPU. The results are listed in Table~\ref{tab:runtime}. 
Our method achieve real-time efficiency, which is suitable for practical deployment. Although GICP can achieve {more robust results} than our network prediction in Table~\ref{tab:eval_kitti} and Table~\ref{tab:eval_apollo}, it is too time-consuming to be directly applied to large-scale LiDAR data. 
\vspace{-1ex}
\section{Conclusions}
\vspace{-1ex}
We present a novel unsupervised LiDAR odometry framework based on the 3D convolutional neural networks. Several 3D self-supervised losses are proposed and uncertainty-aware mechanism is introduced to jointly enable our network to  train in the wild. 
Our method achieve the state-of-art performance on two public datasets. It is worth mentioning that our method can run in real-time and can be combined with other off-the-shelf mapping algorithms for deployment in practical applications.



\clearpage
\acknowledgments{
This work is supported in part by the General Research Fund through the Research Grants Council of Hong Kong under Grants(Nos. CUHK14208417, CUHK14207319), in part by the Hong Kong Innovation and Technology Support Program (No. ITS/312/18FX), in part by CUHK Strategic Fund.
}


\bibliography{final}  

\end{document}